\pdfoutput=1

\documentclass[11pt]{article}

\usepackage[]{emnlp2021}

\usepackage{times}
\usepackage{latexsym}
\usepackage{hyperref}
\usepackage{multirow}
\usepackage{booktabs}
\usepackage{todonotes}
\usepackage{tikz}
\usepackage{siunitx}
\newcommand*\circled[1]{\tikz[baseline=(char.base)]{
            \node[shape=circle,draw,inner sep=2pt] (char) {#1};}}

\newcommand\smalldots{\makebox[0.5em][c]{.\hfil.\hfil.}}

\usepackage[T1]{fontenc}

\usepackage[utf8]{inputenc}

\usepackage{microtype}

\title{How Suitable Are Subword Segmentation Strategies for Translating Non-Concatenative Morphology?}

\author{Chantal Amrhein$^1$ \and Rico Sennrich$^{1,2}$\\
  $^1$Department of Computational Linguistics, University of Zurich\\
  $^2$School of Informatics, University of Edinburgh \\ \medskip
  \texttt{\{amrhein,sennrich\}@cl.uzh.ch}}

\begin{document}
\maketitle
\begin{abstract}
Data-driven subword segmentation has become the default strategy for open-vocabulary machine translation and other NLP tasks, but may not be sufficiently generic for optimal learning of non-concatenative morphology. We design a test suite to evaluate segmentation strategies on different types of morphological phenomena in a controlled, semi-synthetic setting. In our experiments, we compare how well machine translation models trained on subword- and character-level can translate these morphological phenomena. We find that learning to analyse and generate morphologically complex surface representations is still challenging, especially for non-concatenative morphological phenomena like reduplication or vowel harmony and for rare word stems. Based on our results, we recommend that novel text representation strategies be tested on a range of typologically diverse languages to minimise the risk of adopting a strategy that inadvertently disadvantages certain languages.\footnote{Test suite and code available at \url{https://github.com/ZurichNLP/segtest}}
\end{abstract}

\begin{table*}[ht]
    \centering
    \begin{tabular}{ccccc}
     \textbf{Compounding} & \textbf{Circumfixation} & \textbf{Infixation} & \textbf{Vowel Harmony} & \textbf{Reduplication}\\ 
     (German) & (Chickasaw) & (Bontoc) & (Turkish) & (Itza') \\\\ 
     Schild / Kröte & lakna & fikas & \textbf{ü}z\textbf{ü}ld\textbf{ü}n\textbf{ü}z & tz'eek\\
     `shield' / `toad' & `it is yellow' & `strong' & \textbf{ü}z-\textbf{ü}l-d\textbf{ü}-n\textbf{ü}z & `few'\\
     \textbf{Schild}kröte & \textbf{ik}lakn\textbf{o} & f\textbf{um}ikas &  sadden-\small{\texttt{PASS}-\texttt{PAST}-\texttt{2PL}} & \textbf{tz'eek}-tz'eek\\
     `turtle' & `it isn’t yellow' & `to be strong' & `You became sad.' & `very few'\\
      & \small \citep{fromkin2018introduction} & \small\citep{fromkin2018introduction} & \small\citep{goksel2004turkish} & \small\citep{hofling2000itzaj}
    \end{tabular}
    \caption{Examples for the morphological phenomena studied in this paper. The first three are concatenative, the last two non-concatenative.}
    \label{tab:phenomena}
\end{table*}

\section{Introduction}
Data-driven subword-level segmentation of text \citep{sennrich-etal-2016-neural, kudo-2018-subword} is a well-known and widely used text representation strategy in the natural language processing (NLP) community. While subword segmentation largely solves the open vocabulary problem, previous research has shown that models often break down in out-of-domain contexts \citep{el-boukkouri-etal-2020-characterbert}, when encountering spelling errors \citep{belinkov2018synthetic, pruthi-etal-2019-combating}, when translating morphologically-rich languages \citep{ataman-federico-2018-compositional} and in multilingual scenarios \citep{chung-etal-2020-improving, wang-etal-2021-multi}. The reason for this is that even slight deviations from the text seen when learning a segmentation model can result in entirely different segmentations and often aggressively over-segmented text.

Given the rich morphological diversity across natural languages, it is especially interesting to investigate the suitability of subword segmentation to represent different morphological phenomena.
For example, reduplication is a non-concatenative morphological phenomenon\footnote{The distinction between non-concatenative and concatenative morphological phenomena is often a topic of debate in Linguistics. We follow \citet{LieberRochelle2014TOHo}.} that is common across the world's languages, but is marginal in higher-resource European languages,\footnote{See reduplication feature in \href{https://wals.info/feature/27A\#2/28.3/149.2}{WALS} \citep{wals-27}.} which raises the question if the dominant text representation strategies inadvertently disadvantage NLP systems for languages that feature it.

Many types of morphological phenomena (see examples in Table \ref{tab:phenomena}) pose challenges to subword-level models. For concatenative phenomena such as affixes, subword-level segmentations often do not adhere to morpheme boundaries which can hurt the performance of these models. For non-concatenative morphology, it is still unclear to what extent subword-level models can learn to generalise to rare or unseen words. We believe these challenges are exciting opportunities to work on better text representations for cross-lingual NLP but currently, there is a lack of targeted evaluation environments. Most previous work evaluates very specific morphological or morpho-syntactic functions such as number, case, gender or subject-verb agreement \citep{sennrich-2017-grammatical, burlot-yvon-2017-evaluating, warstadt-etal-2020-blimp-benchmark} rather than evaluating how well different \textit{types} of morphological phenomena can be learned.

To address this issue, we design a test suite that can be used to evaluate how well a range of morphological phenomena can be learned with sentence-level sequence-to-sequence models. We focus on the task of neural machine translation (NMT) and evaluate in a semi-synthetic DE$\rightarrow$EN setting, allowing for an automatic evaluation that controls for various confounding factors. In our experiments, we test how well current segmentation strategies on subword- and character-level can learn to translate compounds, circumfixed words, infixed words, vowel harmony and reduplicated words.

Our contributions are the following:

\begin{itemize}
    \item We design an evaluation environment for various types of morphological phenomena that can be used to evaluate future text representation strategies.
    \item We find that non-concatenative morphological phenomena and generalisation to rare word bases are especially challenging to learn with current segmentation strategies.
    \item We show that subword segmentation is less suitable to learn the correct surface form but all segmentation strategies perform well when we represent the morphological phenomena with an abstract token instead.
\end{itemize}

\section{Related Work}

Isolated morphological analysis and reinflection have long been of interest to the NLP community, with yearly shared tasks \citep{kurimo-etal-2010-morpho, vylomova-etal-2020-sigmorphon} that result in dedicated architectures that perform well for many languages \citep{aharoni-goldberg-2017-morphological, makarov-clematide-2018-neural,wu-cotterell-2019-exact,rios-etal-2021-biasing}. However, despite the large morphological diversity of natural languages, many approaches for sentence-level sequence-to-sequence tasks are often only tested on a subset of (morphologically similar) languages and then adopted without much questioning \citep{bender2011achieving}.

One such example is subword-level representation of text \citep{sennrich-etal-2016-neural, kudo-2018-subword} which has contributed greatly to the success of deep learning in various NLP tasks and has become a necessary preprocessing step to train state-of-the-art models \citep{devlin-etal-2019-bert, NEURIPS2020_1457c0d6}. Due to its data-dependent nature, subword segmentation algorithms often produce subword splits that do not adhere to morpheme boundaries which can limit the generalisation to rare or unseen words and can lead to performance loss. Furthermore, it is unclear if models trained with subword segmentation can learn to generalise to non-concatenative morphological phenomena such as reduplication or vowel harmony, even in high-resource settings.

Previous work that evaluated how well morphology or morpho-syntax is captured by sequence-to-sequence models either used contrastive test sets to evaluate whether models assign a higher probability to sentences e.g.\ with correct subject-verb agreement \citep{sennrich-2017-grammatical, marvin-linzen-2018-targeted, warstadt-etal-2020-blimp-benchmark} or probing classifiers to evaluate how well morphological features such as case, number or gender can be predicted from the models' hidden representations \citep{belinkov-etal-2017-neural, vylomova-etal-2017-word, dalvi-etal-2017-understanding, bisazza-tump-2018-lazy, 10.1162/coli_a_00367}. 

Our work is similar to \citet{burlot-yvon-2017-evaluating} who also evaluate morphological competence based on the output of machine translation models rather than probabilities or hidden states. However, instead of morphological features, we are interested in evaluating how well different \textit{types} of morphological phenomena can be learned by sequence-to-sequence models, especially with a focus on their textual representation. Closely related is work by \citet{vania-lopez-2017-characters} who compare language model perplexities for different segmentation strategies on morphologically diverse languages and by \citet{klein-tsarfaty-2020-getting} who show that multilingual BERT \citep{devlin-etal-2019-bert} subwords do not reflect the morphological structure of a non-concatenative language like Hebrew well.

Our setup with synthetic morphological phenomena is similar to work by \citet{wang-eisner-2016-galactic} who generate synthetic treebanks by reordering nodes in existing treebanks for various natural languages. Instead of simply reordering components, we need to apply a more complex preprocessing to generate synthetic morphological phenomena that fit the natural context. We discuss this preprocessing in more detail in Section \ref{sec:preprocessing}.

\section{Morphological Phenomena}

We choose five morphological phenomena which we believe may be hard to learn with subword segmentation strategies. We show a natural language example for each morphological phenomenon in Table \ref{tab:phenomena} and describe them briefly below: \vspace{0.2cm}

\noindent \textbf{Compounding}: A compound is a word composed of more than one free morpheme. Compounding can affect the subword segmentation of the individual components which can make it harder to translate compounds even if the individual parts are seen regularly in the training data.\vspace{0.2cm}

\noindent \textbf{Circumfixation}: A circumfix is an affix that consists of two parts, one added at the start of a word stem, the other at the end. With circumfixation, it is not guaranteed that the subword segmentation adheres to the morpheme boundaries and that the base is segmented in the same way as without any affixation. It may be difficult to learn the correct form for rare or unseen circumfixed words.\vspace{0.2cm}

\noindent \textbf{Infixation}: An infix is an affix inserted inside a word stem. A word with an infix \textit{cannot} be segmented in the same way as without infixation and it is not guaranteed that the segmentation splits the infix into a separate token. Infixation may also be hard to learn for rare or unseen cases.\vspace{0.2cm}

\noindent \textbf{Vowel Harmony}: Vowel harmony is a type of assimilation in which the vowels in a morpheme (e.g.\ an affix) are assimilated to vowels in another morpheme (e.g.\ the word stem). Vowel harmony is a non-concatenative morphological process and it is unclear whether an NMT model trained with subword segmentation can learn to generate the correct vowels for rare or unseen words.\newline

\noindent \textbf{Reduplication}: Reduplication is another non-concatenative morphological process in which the whole word (full reduplication) or a part of a word (partial reduplication) is repeated exactly or with a slight change. In some cases, the repetition can also occur twice (triplication). Reduplication often marks features such as plurality, intensity or size, depending on the language and raises the same generalisation question as vowel harmony.

\section{Segmentation Test Suite}
\label{sec:requirements}
We identify four key requirements for our test suite and address them as follows: 
\begin{itemize}
    \item[1)] \textbf{Understanding and generation:} We want to evaluate both how well morphological phenomena can be analysed and generated on the sentence level. For this reason, we choose machine translation as the context of our evaluation, where morphological phenomena can occur both on the source and the target side.
    \item[2)] \textbf{Automatic Targeted Evaluation:} We want to offer an automatic evaluation to make our test suite independent of resources needed for expensive human evaluation. Morphological phenomena are hard to evaluate automatically in real-data settings where there can be exceptions to morphological rules and ambiguity in how a sentence is translated. Therefore, we decide to evaluate in a semi-synthetic scenario where we have full control over the morphological phenomena and their translations. 
    
    The morphological phenomena should also be evaluated in isolation, i.e. not on the level of BLEU. To achieve this, we do not use naturally occurring morphemes to create our synthetic morphological phenomena. Rather, we generate artificial morphemes that do not occur in our training data otherwise and are distinct between source and target.\footnote{We generate random sequences of consonants and vowels (four to six characters) and check that they do not occur in a subword vocabulary computed on the original training data.} In this way, there are also no cognates among the artificial morphemes we evaluate.
    
    \item[3)] \textbf{Computational Cost:} To keep the computational cost minimal, we decide to insert all synthetic morphological phenomena simultaneously in the training data. Consequently, only a single machine translation model needs to be trained to evaluate a new representation strategy on all morphological phenomena. This reduces the carbon footprint roughly by a factor of five, compared to training models for each phenomenon separately.
    \item[4)] \textbf{Training Data and Vocabulary Size:} The influence of factors such as training data size or vocabulary coverage should be minimised. Therefore, we choose a high-resource data setting where we can easily insert morphological phenomena with varying frequency. If morphological phenomena cannot be learned with ample resources, models will likely perform even worse in real-data, low-resource scenarios.
\end{itemize}

With these requirements in mind, we decide to insert synthetic morphological phenomena in a high-resource DE$\rightarrow$EN translation setting. Per morphological phenomenon of interest, we define a set of patterns that we match in the original sentence and replace with a synthetic morphological phenomenon using the artificial morphemes. The patterns can either be:
\begin{itemize}
    \item A pair of semantically equivalent prepositions, where we synthetically express the prepositional function in either the source or target sentence e.g.\ with an infix.
    \item A pair of semantically equivalent cardinal numbers as noun modifiers, where we synthetically express the cardinality of either the source or target noun e.g.\ with a subsequent token that is subject to vowel harmony.
    \item A pair of semantically equivalent negation particles or intensifiers (such as ``very'') for adjectives, where we synthetically express the modifying function in either the source or target e.g.\ with reduplication of the adjectives.
    \item A pair of semantically equivalent nouns, where we use artificial morphemes to create synthetic compounds in either the source or the target.
\end{itemize}

We choose these types of patterns because they can be expressed morphologically in natural languages. For each type, we select the most frequent pattern pairs in the training and test data. Reduplication often expresses negation or intensification in natural languages, so we assign those patterns to this phenomenon. The remaining patterns that occur frequently enough are mostly prepositional functions. Consequently, circumfixation, infixation and vowel harmony are both assigned prepositional patterns. A full overview of all pattern pairs and artificial morphemes for each morphological phenomenon is listed in Appendix \ref{sec:appendixb}.

\section{Experimental Setup}

\subsection{Data Sources}
Our training data consists of $\sim$ 4.6M parallel sentences from the WMT16 shared task training data \citep{bojar-etal-2016-findings}. For development, we take the test set from WMT15 ($\sim$ 2k parallel sentences) and for testing, the test sets from all other years of the shared task ($\sim$ 28k parallel sentences).

\begin{table*}[ht]
    \centering
    
    \begin{tabular}{cccc}
        \multirow{3}{*}{\small Compound} & \texttt{O} & \small Die \textbf{Räume} seien vorhanden. & \small The \textbf{premises} are available.\\
         & \texttt{S} & \small Die \textbf{Sonaräume} seien vorhanden. & \small The \textbf{bico premises} are available.\\
         & \texttt{A} &  \small Die \textbf{Räume @COMPOUND\_1@} seien vorhanden. & \small The \textbf{wuze premises} are available. \\ \addlinespace
         
         \multirow{3}{*}{\small Circumfix} & \texttt{O} & \small Das sind gute Nachrichten \textbf{für die Stadt}. & \small That is good news \textbf{for the city}.\\
        & \texttt{S} & \small Das sind gute Nachrichten \textbf{wofi die Stadt}. & \small That is good news \textbf{the jebcityfet}.\\
        & \texttt{A} & \small Das sind gute Nachrichten \textbf{fuge die Stadt}. & \small That is good news \textbf{the city @CIRCUMFIX\_1@}.\\ \addlinespace
        
         \multirow{3}{*}{\small Infix} & \texttt{O} &  \small Er schimpfte \textbf{bei der Kritik}, sicher. & \small He chafed \textbf{at the criticism}, sure. \\
        & \texttt{S} & \small Er schimpfte \textbf{der Kryadeyitik}, sicher. & \small He chafed \textbf{numime the criticism}, sure.\\
        & \texttt{A} & \small Er schimpfte \textbf{der Kritik @INFIX\_4@}, sicher. & \small He chafed \textbf{jigaq the criticism}, sure.  \\ \addlinespace
        
         \multirow{2}{*}{\small Vowel} & \texttt{O} & \small Das waren gleich \textbf{zwei Fehler}! & \small Those were \textbf{two errors}!\\
        \multirow{2}{*}{\small Harmony} & \texttt{S} & \small Das waren gleich \textbf{zoged Fehler}! & \small Those were \textbf{errors bepor}!\\
        & \texttt{A} & \small Das waren gleich \textbf{gapu Fehler}! & \small Those were \textbf{errors @VOWEL\_HARMONY\_2@}!\\ \addlinespace
        
         \multirow{3}{*}{\small Redupl.} & \texttt{O} & \small Das ist \textbf{nicht gefährlich}. & \small This is \textbf{not dangerous}.\\
        & \texttt{S} & \small Das ist \textbf{gija gefährlich}. & \small This is \textbf{dangerousdangerous}. \\
        & \texttt{A} & \small Das ist \textbf{jufo gefährlich}. & \small This is \textbf{dangerous @FULL\_REDUPLICATION@}. \\ 
    \end{tabular}
    \caption{Surface (\texttt{S}) and abstract examples (\texttt{A}) for all morphological phenomena and the original sentences (\texttt{O}). German source sentences on the left, English target sentences on the right. The modified token spans are marked in bold.}
    \label{tab:synthetic}
\end{table*}

\subsection{Preprocessing}
\label{sec:preprocessing}
\textbf{Word Alignment:} We first word-align our parallel sentences. Word alignments are used to ensure the morphological phenomena are inserted in the corresponding source and target tokens. We use \texttt{eflomal} \citep{Ostling2016efmaral} to learn the word alignment.\newline

\noindent \textbf{Parsing:} We also parse our data to be able to write more specific matching rules. We use pretrained \texttt{spaCy} \citep{spacy} parsers\footnote{English: en\_core\_web\_md, German: de\_core\_news\_md} and the \texttt{spacy\_conll} library \footnote{\url{github.com/BramVanroy/spacy\_conll}} to create \href{https://universaldependencies.org/format.html}{CoNLL-U format}. Through this format, we also have access to part-of-speech (POS) tags and the lemmas of the tokens.

\subsection{Inserting Morphological Phenomena} 
To insert the synthetic morphological phenomena, we first check if the corresponding pattern pair (prepositions, cardinal ``two'' or modifier for adjectives) occurs in the source and target sentence. If this is the case, we check whether the patterns are aligned and extract the tokens where we want to insert the synthetic morphological phenomena. For prepositional functions, this is the noun of the prepositional phrase, for the cardinal ``two'', this is the noun that the cardinal modifies and for adjective modifying functions, this is the adjective following the modifier. We find these tokens using the information from the POS-tags and the dependency parse and check that the tokens are also aligned translations of each other.\footnote{98.4\% synthetic phenomena were introduced correctly in a manual evaluation of 200 random sentence pairs per morphological phenomenon, despite automatic alignment and parsing.}

In one sentence (either the source or the target), we use an artificial morpheme to create the synthetic morphological phenomenon and delete the preposition, cardinal number or adjective modifier. In the other sentence, we replace the preposition, cardinal number or modifier with another, isolated artificial morpheme. For compounds, we concatenate an artificial morpheme with a random noun in the source and introduce another, isolated artificial morpheme before the corresponding translation of that noun in the target. We never insert synthetic morphological phenomena on both sides simultaneously, i.e. one of the artificial morphemes in each pair is always isolated. The artificial morphemes are also unique for each pattern pair to minimise interference between them. Some examples for the resulting sentences can be seen in Table \ref{tab:synthetic}.

To better evaluate how hard it is for a model to learn a specific morphological phenomenon, we also create sentence pairs with an abstract representation of the morphological phenomenon as a control, similar to \citet{tamchyna-etal-2017-modeling}. Instead of modifying the surface form, this abstract representation is simply an additional token that is used to indicate that the preceding token is subject to a specific morphological phenomenon. Results with this abstract representation act as an upper bound in our evaluation setup, indicating how well a model could learn a morphological phenomenon if it had access to an oracle to either analyse or produce the correct surface form. 

When all modified sentences are added to the original training data we obtain a total training set with $\sim$ 5.6M sentence pairs. We add the modified sentences instead of replacing original sentences so that the use of our test suite does not impair the translation quality on real text.\footnote{We compare dev BLEU per checkpoint for models without and with added synthetic morphological phenomena and observe only an average absolute difference of ~0.25 with 32k merges and ~0.1 with 500 merges.} This way, future work could include our test suite training data without needing to train separate models for measuring general performance, thus minimising effort and carbon footprint. For testing, we only choose the sentence pairs where we inserted morphological phenomena. Depending on the pattern pair, we have between 50 and 700 test examples each. The exact numbers are presented in Appendix \ref{sec:appendixb} and we also present results with synthetically balanced test sets in Appendix \ref{sec:appendixd}.

\subsection{Model Description}
We train four neural machine translation models on our modified training data:
\begin{itemize}
    \item[\circled{1}] A subword-level BPE model \cite{sennrich-etal-2016-neural} with 32k merge-operations as a baseline and representation of current state-of-the-art models.
    \item[\circled{2}] A subword-level BPE model with 32k merge-operations, trained with BPE-dropout \citep{provilkov-etal-2020-bpe}. With BPE-dropout, the training data is resegmented after every epoch and at each merge step, some merges are randomly dropped. Like other dropout methods \citep{JMLR:v15:srivastava14a, zhou-etal-2020-scheduled}, BPE-dropout has a regularising effect and allows the model to generalise better to text segmented into smaller units. We also expect it to help generalisation across different occurrences of the same morpheme that would be segmented differently with deterministic subword segmentation.
    \item[\circled{3}] A subword-level BPE model with 500 merges, trained with BPE-dropout. This model learns on much smaller subword units although not completely on character-level and is used as a parent model for the next model.
    \item[\circled{4}] A character-level model which is finetuned on the model with 500 merge-operations. This finetuning strategy allows training reasonably well-performing character-level models without the need for very deep architectures \citep{libovicky-fraser-2020-towards}.
\end{itemize} 

We train Transformer Base machine translation models \citep{NIPS2017_7181} with the \texttt{nematus}\footnote{\url{github.com/EdinburghNLP/nematus}} \citep{sennrich-etal-2017-nematus} framework. We train the first three models for 700k updates and choose the best checkpoint based on the BLEU score. This is evaluated on a dev set without synthetic morphological phenomena using SacreBLEU\footnote{BLEU+case.mixed+lang.de-en+numrefs.1+smooth.exp+tok.13a+version.1.4.2} \citep{post-2018-call}. For the character-level model, we start the finetuning from the best checkpoint in the first 400k updates of the subword model with 500 merges. The character-level model is then finetuned for an additional 550k updates and we choose the best checkpoint based on BLEU as for the other models.

Our subword vocabularies are computed with byte pair encoding \citep{sennrich-etal-2016-neural} using the  SentencePiece implementation \citep{kudo-richardson-2018-sentencepiece}. We use a character coverage of 0.9999 to ensure the vocabulary for the model with 500 subword segmentation operations does not consist of virtually only single characters. With this restriction, the vocabulary of our character-level model consists of  246 single characters plus three reserved tokens used by the NMT model and the 25 morphological tokens used for the abstract representations of the morphological phenomena. We provide more details on hyperparameters and computing environment in Appendix \ref{sec:appendix}.

\subsection{Evaluation}
Since we use artificial morphemes to mark the morphological phenomena, we can evaluate if the correct artificial morpheme is produced rather than comparing to a reference. For phenomena occurring on the source side, we simply need to check whether the correct artificial morpheme that e.g.\ replaced a preposition or intensity marker occurs in the model's output sentence. On the target side, the evaluation is a bit more complex. For circumfixation, we check if a token exists that is circumfixed with the correct artificial morphemes. For infixation, we check if there is a token that is infixed with the correct artificial morpheme. For vowel harmony, we check if the correct consonant triple occurs in the output sentence and whether the vowels between the consonants agree with the last two vowels of the previous token. For reduplication on the target side (full reduplication), we check if there is a fully repeated token in the output sentence. We do not evaluate whether the base of the phenomena matches the reference since only the translation of the artificial morphemes is guaranteed to be unambiguous in our training data. With this evaluation setup, we can compute the accuracy over all test sentences that contain a morphological phenomenon.

\begin{table}[]
    \centering
    \small
    \begin{tabular}{cccccc}
        & bpe32k & bpe-d32k & bpe-d500  & char\\ 
        \cmidrule(lr){2-2}  \cmidrule(lr){3-3}  \cmidrule(lr){4-4}   \cmidrule(lr){5-5} \addlinespace
        dev & \textbf{31.44} &  31.03 & 30.14 & 29.78\\\addlinespace
        test & \textbf{30.19} &  30.02 &  28.82 & 28.66\\\addlinespace
    \end{tabular}
    \caption{BLEU scores on the development and test set (without morphological phenomena).}
    \label{tab:bleu}
\end{table}

\section{Results}

\begin{table*}[t]
    \centering
    \begin{tabular}{crcS[table-number-alignment = right]cccccccc}
    & & &  & \multicolumn{4}{c}{\textbf{Surface Representation}} & \textbf{Abstract}\\
    \cmidrule(lr){5-8} \cmidrule(lr){9-9} 
    & & \textbf{Side} & \textbf{Train Freq.} &bpe32k & bpe-d32k & bpe-d500  & char & bpe32k\\
        \cmidrule(lr){3-3}   \cmidrule(lr){4-4}  \cmidrule(lr){5-5}  \cmidrule(lr){6-6}  \cmidrule(lr){7-7}  \cmidrule(lr){8-8} \cmidrule(lr){9-9} \addlinespace
    \multirow{5}{*}{\textbf{Compounding}} & \#9 & src & 27 & \phantom{0}0.0 & \phantom{0}0.0 & \phantom{0}0.0 & \phantom{0}0.0 & \phantom{0}0.0\\ 
    & \#7 & src & 67 & \colorbox[HTML]{F17779}{46.1} & \phantom{0}0.0 & \colorbox[HTML]{F9CBD0}{\textbf{83.8}} & \phantom{0}0.0 & 95.1\\
    & \#5 & src & 238 & \colorbox[HTML]{5CC35C}{\textbf{98.1}} & \colorbox[HTML]{5CC35C}{97.6} & \colorbox[HTML]{5CC35C}{96.2} & \colorbox[HTML]{5CC35C}{97.0} & 98.1\\  
    & \#3 & src & 522 & \colorbox[HTML]{5CC35C}{\textbf{98.9}} & \colorbox[HTML]{5CC35C}{98.4} & \colorbox[HTML]{5CC35C}{97.3} & \colorbox[HTML]{5CC35C}{96.5} & 98.1 \\
    & \#1 & src & 1095 & \colorbox[HTML]{5CC35C}{96.2} & \colorbox[HTML]{5CC35C}{\textbf{97.8}} & \colorbox[HTML]{5CC35C}{97.3} & \colorbox[HTML]{5CC35C}{97.0} & 96.5 &\\ \addlinespace \addlinespace
    
    
    \multirow{4}{*}{\textbf{Circumfixation}} & \#4 & src & 11718 & \colorbox[HTML]{5CC35C}{\textbf{97.9}} & \colorbox[HTML]{5CC35C}{\textbf{97.9}} & \colorbox[HTML]{5CC35C}{\textbf{97.9}} & \colorbox[HTML]{5CC35C}{95.9} & 97.9\\ 
    & \#2 & src & 26007 & \colorbox[HTML]{5CC35C}{\textbf{100}} & \colorbox[HTML]{5CC35C}{98.0} & \colorbox[HTML]{5CC35C}{98.0} & \colorbox[HTML]{5CC35C}{99.2} & 99.6 \\  \addlinespace
    & \#3 & trg & 21372 & \colorbox[HTML]{5CC35C}{97.3} & \colorbox[HTML]{5CC35C}{\textbf{100}} & \colorbox[HTML]{5CC35C}{97.3} & \colorbox[HTML]{5CC35C}{96.4} & 100\\
    & \#1 & trg & 122017 & \colorbox[HTML]{5CC35C}{96.3} & \colorbox[HTML]{5CC35C}{\textbf{99.0}} & \colorbox[HTML]{5CC35C}{\textbf{99.0}} & \colorbox[HTML]{5CC35C}{97.4} & 99.4 \\ \addlinespace \addlinespace
    \multirow{4}{*}{\textbf{Infixation}} & \#4 & src & 3796 & \colorbox[HTML]{5CC35C}{98.9} & \colorbox[HTML]{5CC35C}{98.9} & \colorbox[HTML]{5CC35C}{96.7} & \colorbox[HTML]{5CC35C}{\textbf{100}} & 100 \\  
    & \#3 & src & 15540 & \colorbox[HTML]{5CC35C}{98.5} & \colorbox[HTML]{5CC35C}{96.4} & \colorbox[HTML]{5CC35C}{\textbf{99.3}} & \colorbox[HTML]{5CC35C}{97.1} & 97.8 \\ \addlinespace
    & \#2 & trg & 47102 & \colorbox[HTML]{5CC35C}{97.2} & \colorbox[HTML]{5CC35C}{\textbf{98.6}} & \colorbox[HTML]{5CC35C}{\textbf{98.6}} & \colorbox[HTML]{5CC35C}{97.2} & 99.6\\ 
    & \#1 & trg & 116868 & \colorbox[HTML]{5CC35C}{97.9} & \colorbox[HTML]{5CC35C}{\textbf{98.9}} & \colorbox[HTML]{5CC35C}{98.3} & \colorbox[HTML]{5CC35C}{98.7} &99.6\\ \addlinespace \addlinespace
    \multirow{4}{*}{\textbf{Vowel Harmony}} & \#3 & src & 8636 & \colorbox[HTML]{5CC35C}{98.9} & \colorbox[HTML]{5CC35C}{\textbf{99.4}} & \colorbox[HTML]{5CC35C}{97.7} & \colorbox[HTML]{5CC35C}{98.3} &98.9\\ \addlinespace
    & \#4 & trg & 7037 & \colorbox[HTML]{F17779}{70.4} & \colorbox[HTML]{F9CBD0}{80.0} & \colorbox[HTML]{B2EAB1}{\textbf{91.3}} & \colorbox[HTML]{B2EAB1}{90.4} & 99.1\\ 
    & \#2 & trg & 29048 & \colorbox[HTML]{F17779}{78.9} & \colorbox[HTML]{F9CBD0}{82.7} & \colorbox[HTML]{B2EAB1}{\textbf{93.5}} & \colorbox[HTML]{B2EAB1}{92.9} & 99.4\\ 
    & \#1 & trg & 133082 & \colorbox[HTML]{F9CBD0}{82.0} & \colorbox[HTML]{F9CBD0}{87.9} & \colorbox[HTML]{B2EAB1}{93.8} & \colorbox[HTML]{B2EAB1}{\textbf{94.4}} & 99.4 \\  \addlinespace \addlinespace
    \multirow{3}{*}{\textbf{Reduplication}} & Triple & src & 106 & \phantom{0}0.0	& \phantom{0}0.0 & \phantom{0}0.0 & \phantom{0}0.0 & 99.2 \\ 
    & Partial & src & 34783 &  \colorbox[HTML]{B2EAB1}{94.2} & \colorbox[HTML]{5CC35C}{95.0} & \colorbox[HTML]{5CC35C}{\textbf{95.9}} & \colorbox[HTML]{5CC35C}{95.0} & 99.2 \\  \addlinespace
    & Full & trg & 9664 & \colorbox[HTML]{F17779}{72.0} & \colorbox[HTML]{F9CBD0}{84.0} & \colorbox[HTML]{B2EAB1}{\textbf{94.0}}  & \colorbox[HTML]{B2EAB1}{90.0} & 98.0
    \end{tabular}
    \caption{Accuracy (in \%) of the four models for each of the morphological pattern pairs. Best results for surface representation  are marked in bold. $\geq$95\% dark green, $\geq$90\% light green,  $\geq$80\% light red,  $<$80\% dark red (best viewed in colour). Patterns ordered by src / trg side, then by frequency.}
    \label{tab:coarse_results}
\end{table*}

\subsection{Translation Quality}
First, we show a quick overview of the translation quality of our models. Table \ref{tab:bleu} shows the BLEU scores on the original dev and test sets without the inserted morphological phenomena. While the subword model with 32k merges without BPE-dropout performs best, the model with BPE-dropout does not perform much worse on the test set. Training NMT models with smaller units decreases the translation quality by $\sim$ 1.5 BLEU for the subword model with 500 merges and the character-level model compared to the best model.

\subsection{Concatenative Morphology}
An evaluation on the level of BLEU does not offer any insight into how well these models can handle the morphological phenomena we are interested in. Table \ref{tab:coarse_results} shows the accuracy results on our test suite for each of the morphological pattern pairs. For compounding, it is interesting to see how the accuracy changes with increasing frequency of the patterns in the training data. Even with an abstract representation (last column), it takes around 70 training examples for the subword-level models with 32k merges to learn to translate the phenomenon correctly. 

The results for circumfixation and infixation further show that concatenative morphological phenomena can be learned rather well by all models on both sides and with both the abstract and surface representations. The results of the other models with the abstract representation are comparable to the subword model with 32k merges.

\subsection{Non-Concatenative Morphology}
The most interesting results can be seen for the two non-concatenative morphological phenomena: vowel harmony and reduplication. First, there is a clear gap between the accuracy with the abstract representation and the surface representation for the subword-level model with 32k merges. The only exception is vowel harmony pattern \#3 where the vowel harmony occurs on the source side and does not need to be generated by the model. Second, we can again see an effect of the frequency of the pattern pairs. For vowel harmony, the accuracy drops significantly the rarer a pattern pair is. This is more prominent in the subword models with 32k splits. Similarly, for reduplication, while partial reduplication which occurs $\sim$35k times in the training data can be learned to some extent, none of the models can learn to translate triplication correctly which is only seen 106 times. These results indicate that non-concatenative morphological phenomena will be even harder to learn in real-life scenarios, where we often encounter low-resource settings and  more ambiguity in the translations.

For the non-concatenative morphological phenomena, we can see a considerable benefit from translating with smaller units. Even simply using BPE-dropout at training time can give a boost of up to 20\% in accuracy. Given these results, we support the recommendation by \citet{wang-etal-2021-multi} that BPE-dropout should become the default for training sequence-to-sequence models.

\begin{figure}[!t]
    \centering
    \includegraphics[width=0.47\textwidth]{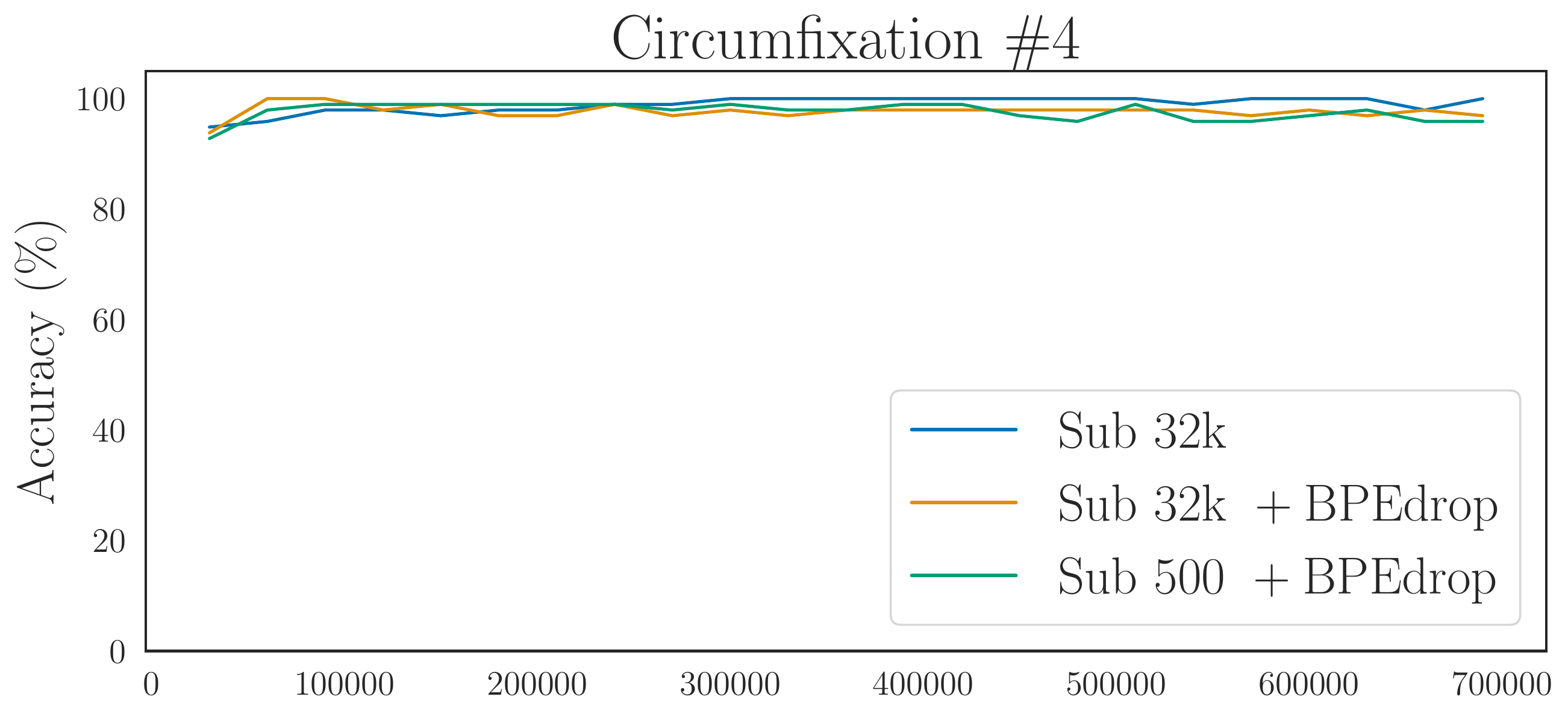} \vspace{0.1cm}
    
    \includegraphics[width=0.47\textwidth]{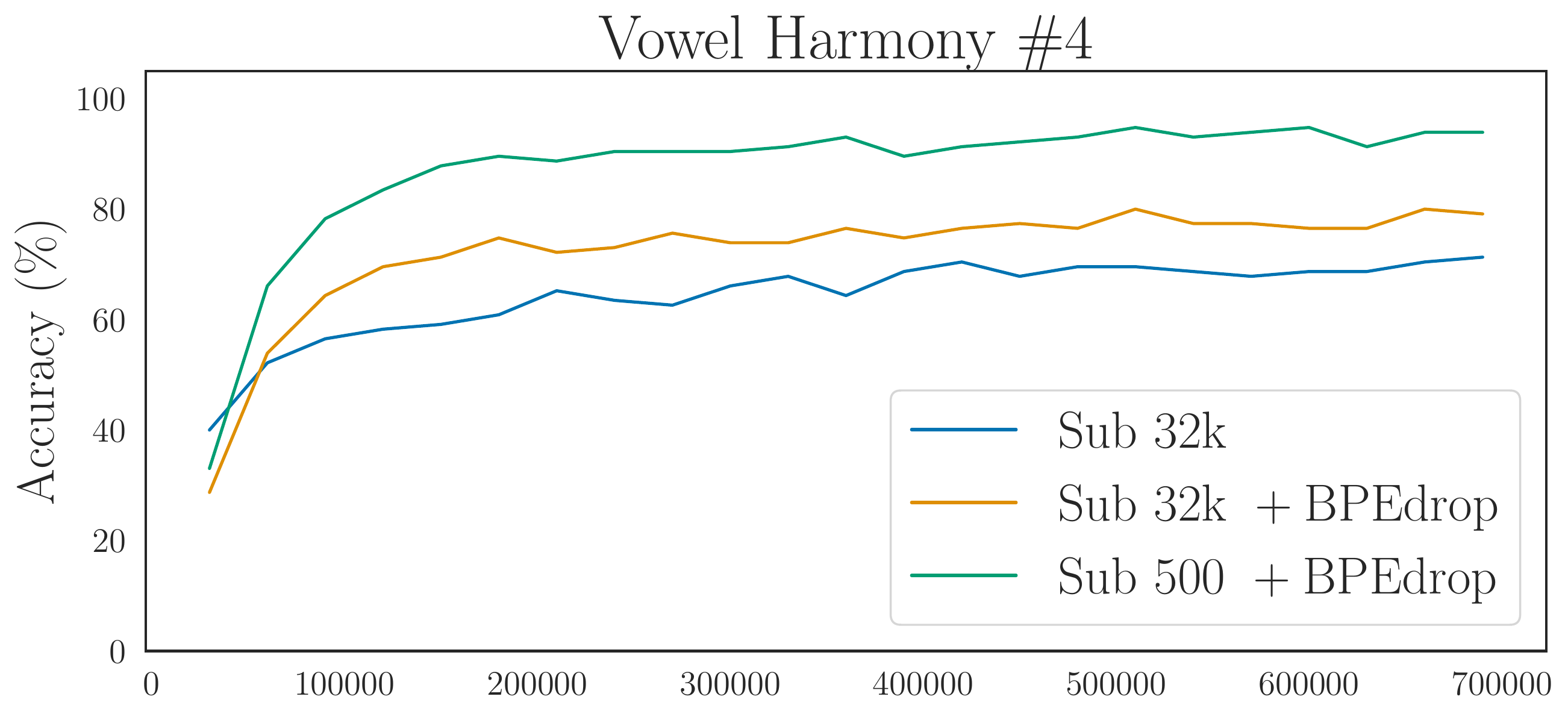} \vspace{0.1cm}
    
    \includegraphics[width=0.47\textwidth]{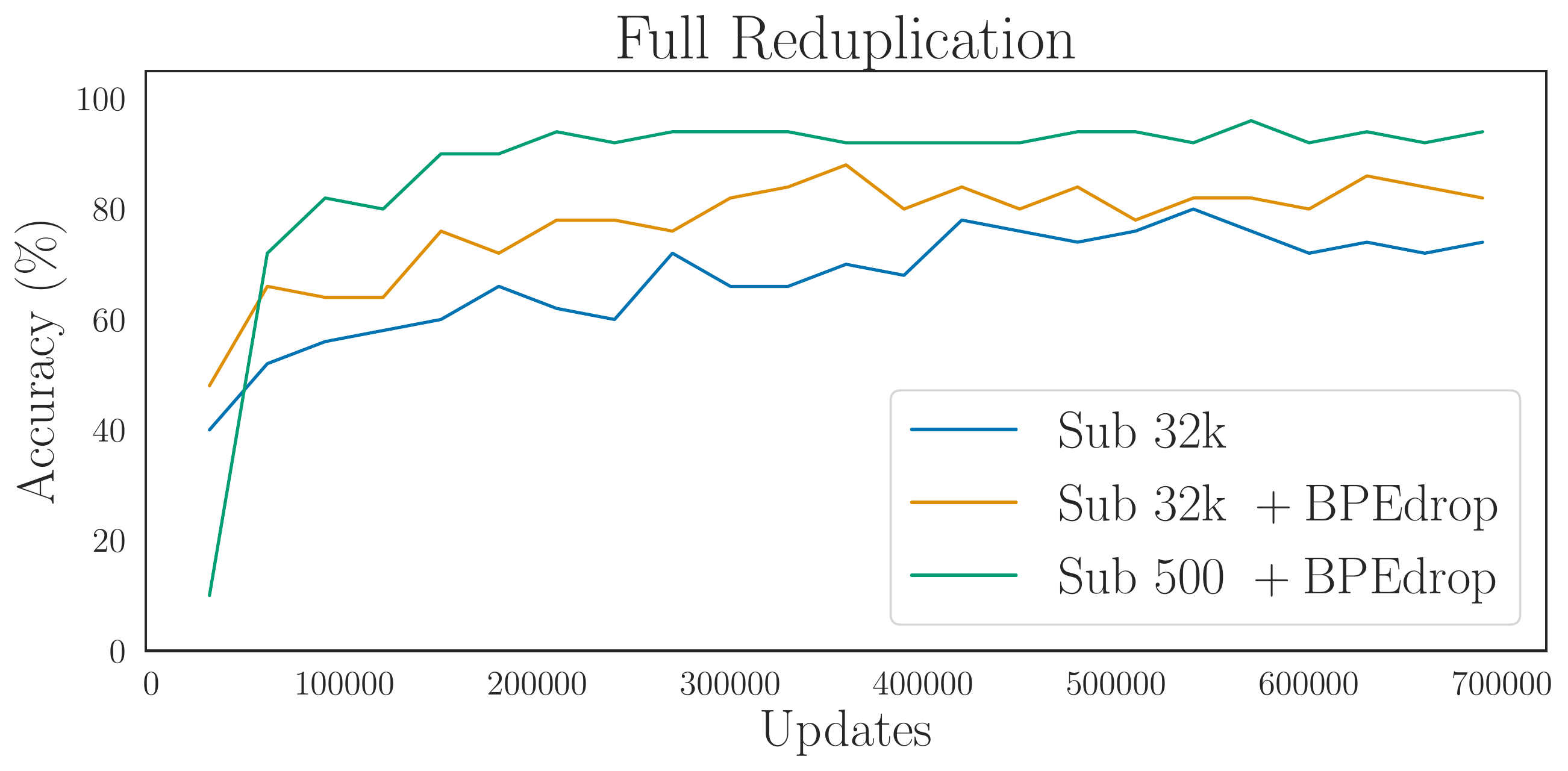}
    \caption{Accuracy of one pattern pair per morphological phenomenon over time.}
    \label{fig:over_time}
\end{figure}

\section{Analysis and Discussion}
\subsection{Learning Over Time}
It is interesting to see how the models learn to translate the different morphological patterns over time. Figure \ref{fig:over_time} shows the training curves for the first three models\footnote{We do not show training curves for the character-level model because it was not trained from scratch.} on a circumfixation, vowel harmony and reduplication pattern pair. The pattern frequencies in the training data are comparable, occurring 11718, 7037 and 9664 times respectively.

While circumfixation is learned almost perfectly after the first few checkpoints by all models, we can see that reduplication and especially vowel harmony are learned much more slowly. For the latter two phenomena, we can also see that the differences between the two models are much more pronounced, e.g. for vowel harmony the subword model with 500 merges continuously outperforms the other two models. These plots also still show an improving tendency at 700k steps, so longer training times may be beneficial for learning non-concatenative morphology in NMT.

\subsection{Source Frequency During Training}

\begin{figure}[!t]
    \centering
    \includegraphics[width=0.48\textwidth]{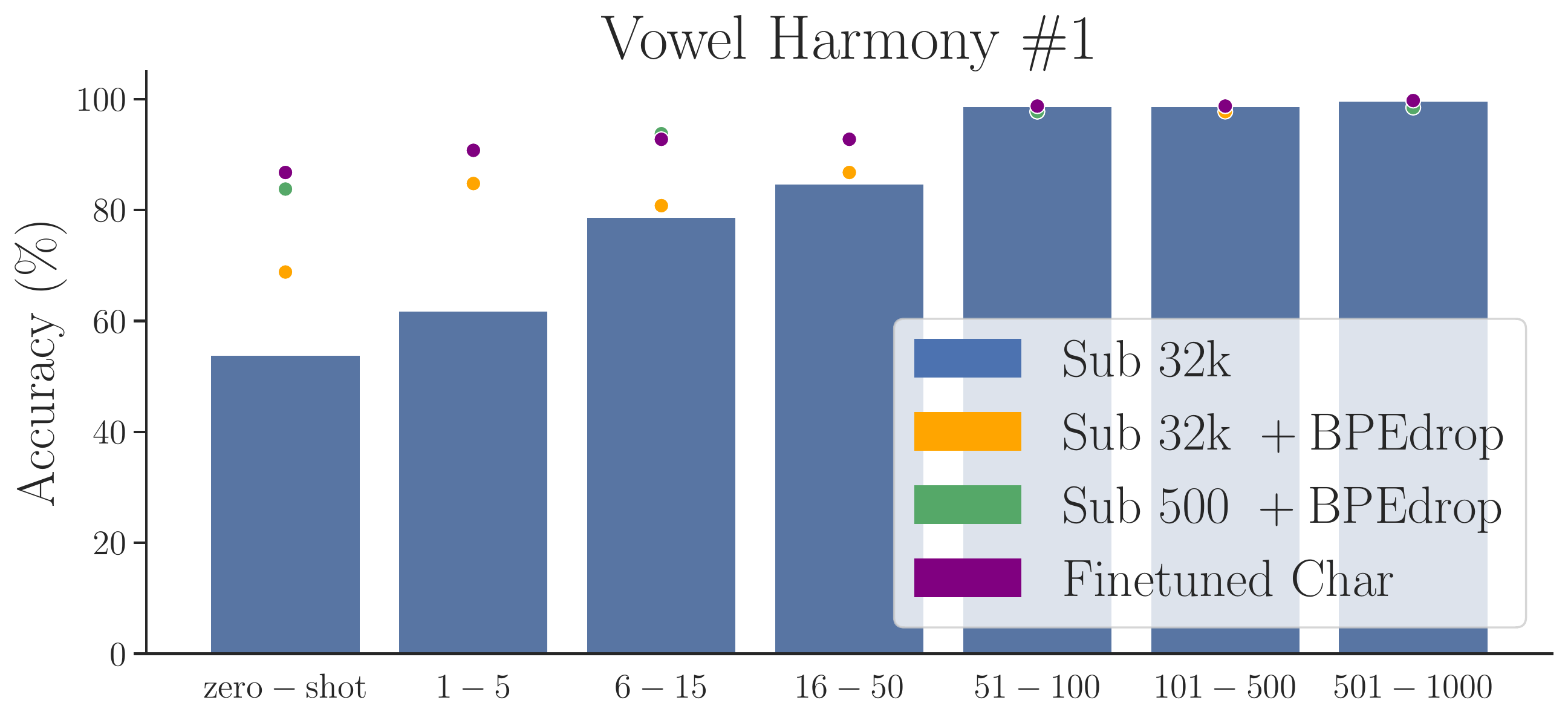} 
    \caption{Accuracy on different frequency buckets according to how often the source token has been seen with the morphological pattern during training.}
    \label{fig:src_freq}
\end{figure}

We perform a more fine-grained evaluation and bucket test sentences according to how often the modified token in the source occurred with the specific morphological pattern in the training data. Figure \ref{fig:src_freq} shows that the main benefits from using models with smaller units come from the better generalisation to unseen or very rare modified tokens. This finding suggests that character-level models may outperform subword-level models to an even greater extent in real-data low-resource settings.

\subsection{Error Types}
\label{sec:error_types}
We perform a manual analysis of up to 50 sentences per pattern pair where the morphological phenomenon was not translated with the correct artificial morphemes. We summarise the most interesting findings here and list the full results in Appendix \ref{sec:appendixc}. For vowel harmony on the target side, we find that all wrong translations are due to vowels that do not match with the previous token. For full reduplication on the target side, we see an interesting effect with the models trained with 32k merges. Instead of reduplicating an adjective such as ``compulsorycompulsory'', these models often concatenate two words that are similar in meaning such as ``mandatorycompulsory''. This effect disappears when training on smaller units.

For rare compounds, the models trained with 32k merges often either copy the whole source compound or the artificial morpheme to the target side. This happens less often with the models trained on smaller units but instead, these models only start to translate the first characters of the artificial source morpheme or hallucinate real words with similar orthography, e.g. ``kidnapping rights'' or ``kidney inhabitants'' instead of ``kixaka rights'' and ``kixaka inhabitants'' respectively. 

\subsection{How Realistic Are Our Results?}
We note that our results should not be taken as evidence that current NMT models can perfectly translate concatenative morphology. Generally, we expect that our controlled setting - where there is a one-to-one correspondence between artificial morphemes - is an idealised scenario and that models likely perform worse in real-life settings with more ambiguity and noise. However, our results do show a clear gap between the models' competence for non-concatenative and concatenative morphology. Considering this performance gap and our reasons for evaluating in a semi-synthetic setup (see Section \ref{sec:requirements}), we think that our test suite offers a targeted way to compare how well novel text representation strategies can learn non-concatenative phenomena. 

For vowel harmony, there is one factor in our setting that may slightly increase its difficulty: only a few patterns in our data set exhibit vowel harmony. This might make it harder for the model to learn to extract the relevant information (i.e.\ the vowels in word stems) than if all suffixes in a language followed vowel harmony rules. Note however, that the frequencies of the individual patterns in our test suite are realistic. The top 50 nominal inflectional suffixes in Turkish - a language that shows extensive vowel harmony in suffixes - range from ~13’000’000 at rank 1 to ~30’000 at rank 50 \citep{Aksan2016AFD}. These frequencies were counted in a corpus with 50M tokens. Our two more frequent pattern pairs lie in this range and our two less frequent ones capture the long tail of suffixes and more accurately predict expected results in low-resource scenarios (likely less than 50M tokens).

\section{Conclusion}
We develop a test suite to evaluate how well various types of morphological phenomena can be translated in NMT. We show that the choice of segmentation strategy can have a considerable influence on the performance, especially for non-concatenative phenomena such as reduplication and vowel harmony. Our results with current segmentation strategies show a) that there is potential for more work on text representation strategies, b) that abstract representations may be a helpful source of information, especially for languages with non-concatenative morphology (if reliable tools for morphological analysis and generation are available) and c) that BPE-dropout should be adopted in state-of-the-art models since it improves learning non-concatenative morphology. Based on our results, we recommend that novel approaches in NLP always be tested on a range of typologically diverse languages that cover different types of morphological phenomena.

In the future, we are interested in evaluating a wider variety of text representation strategies, including tokenisation-free input such as CANINE \citep{clark2021canine} or visual text representations \citep{salesky2021robust}, although these are limited to the source side. We would also like to investigate the effects of out-of-domain contexts where we expect more rare word stems.

\section*{Acknowledgements}
We thank our colleagues Annette Rios, Phillip Ströbel
and the anonymous reviewers for their helpful feedback. This work was funded by the Swiss National Science Foundation (project MUTAMUR;
no. 176727).

\bibliography{anthology,custom}
\bibliographystyle{acl_natbib}

\newpage

\appendix

\section{Appendix}
\subsection{Additional Model Details}
\label{sec:appendix}
We train Transformer Base machine translation models \citep{NIPS2017_7181} with 6 encoder layers, 6
decoder layers, 8 attention heads, an embedding and hidden state dimension of
512 and a feed-forward network dimension of 2048. We regularise our models with a dropout of 0.1 for the embeddings, the residual connections, in the feed-forward sub-layers and for the attention weights. For BPE-dropout \citep{provilkov-etal-2020-bpe}, we also use a dropout probability of 0.1 during training. 

We apply exponential smoothing of 0.0001 and label smoothing of 0.1. We tie both our encoder and decoder input embeddings as well as the decoder input and output embeddings \citep{press-wolf-2017-using}. The subword models with 32k merges are trained with a maximum sequence length of 200 tokens, the subword model with 500 merges with a maximum sequence length of 500 and the character-level model with a maximum sequence length of 1,000 tokens.

For optimisation, we use Adam \citep{DBLP:journals/corr/KingmaB14} with standard hyperparameters and a learning rate of  0.0001. We follow the Transformer learning schedule described in \citep{NIPS2017_7181} with a linear warmup over 4,000 steps. For finetuning, we use a constant learning rate of 0.001. Our token batch size is set to 16,348 and we train on 4 NVIDIA Tesla V100 GPUs. All models were trained using the implementation provided in \texttt{nematus} \citep{sennrich-etal-2017-nematus} allowing early stopping on a development set with patience 5.

\begin{table*}[]
    \small
    \centering
    \begin{tabular}{rrrrrrcc}
        & \multicolumn{7}{c}{\textbf{Compounding}} \\ 
        \cmidrule(lr){2-8} 
        & \multicolumn{1}{c}{\textbf{SRC pattern}} & \multicolumn{1}{c}{\textbf{TRG pattern}} & \multicolumn{1}{c}{\textbf{SRC artificial}} & \multicolumn{1}{c}{\textbf{TRG artificial}} & \multicolumn{1}{c}{\textbf{train}} & \multicolumn{1}{c}{\textbf{test}} & \multicolumn{1}{c}{\textbf{aug\_test}}  \\ 
        \cmidrule(lr){2-2}  \cmidrule(lr){3-3} \cmidrule(lr){4-4} \cmidrule(lr){5-5} \cmidrule(lr){6-6} \cmidrule(lr){7-7} \cmidrule(lr){8-8}
        \#1 & random \texttt{NOUN} & aligned \texttt{NOUN} & Sona$+$\texttt{NOUN} & bico \texttt{NOUN} & 1095 & 371 & 400\\
        \#3 & random \texttt{NOUN} & aligned \texttt{NOUN} & Suyi$+$\texttt{NOUN} & saqo \texttt{NOUN} & 522 & 371 & 400\\
        \#5 & random \texttt{NOUN} & aligned \texttt{NOUN} & Zarumo$+$\texttt{NOUN} & vazaga \texttt{NOUN} & 238 & 371 & 400\\
        \#7 & random \texttt{NOUN} & aligned \texttt{NOUN} & Necib$+$\texttt{NOUN} & kixaka \texttt{NOUN} & 67 & 371 & 400 \\
        \#9 & random \texttt{NOUN} & aligned \texttt{NOUN} & Dawida$+$\texttt{NOUN} & nonujo \texttt{NOUN} & 27 & 371 & 400 \\ \addlinespace \addlinespace
        & \multicolumn{7}{c}{\textbf{Circumfixation}} \\
         \cmidrule(lr){2-8} 
        & \multicolumn{1}{c}{\textbf{SRC pattern}} & \multicolumn{1}{c}{\textbf{TRG pattern}} & \multicolumn{1}{c}{\textbf{SRC artificial}} & \multicolumn{1}{c}{\textbf{TRG artificial}} & \multicolumn{1}{c}{\textbf{train}} & \multicolumn{1}{c}{\textbf{test}} & \multicolumn{1}{c}{\textbf{aug\_test}}  \\ 
        \cmidrule(lr){2-2}  \cmidrule(lr){3-3} \cmidrule(lr){4-4} \cmidrule(lr){5-5} \cmidrule(lr){6-6} \cmidrule(lr){7-7} \cmidrule(lr){8-8}
        \#1 & für [\smalldots] \texttt{NOUN} &  for [\smalldots] \texttt{NOUN} & wofi [\smalldots] \texttt{NOUN} & jeb$+$\texttt{NOUN}$+$fet & 122 017 & 493 & 700\\
        \#2 & aus [\smalldots] \texttt{NOUN}  & from [\smalldots] \texttt{NOUN}  & Kur$+$\texttt{NOUN}$+$maz & quroc  [\smalldots] \texttt{NOUN} & 26 007 & 256 & 600\\
        \#3 &  zwischen [\smalldots] \texttt{NOUN} & between [\smalldots] \texttt{NOUN}  & seyet  [\smalldots] \texttt{NOUN} & nuw$+$\texttt{NOUN}$+$daf & 21 372 & 110 & 700 \\
        \#4 & durch [\smalldots] \texttt{NOUN}  & through [\smalldots] \texttt{NOUN}  & Rül$+$\texttt{NOUN}$+$bos & sudizu [\smalldots] \texttt{NOUN} & 11718 & 97 & 600 \\ \addlinespace \addlinespace
        & \multicolumn{7}{c}{\textbf{Infixation}} \\
         \cmidrule(lr){2-8} 
        & \multicolumn{1}{c}{\textbf{SRC pattern}} & \multicolumn{1}{c}{\textbf{TRG pattern}} & \multicolumn{1}{c}{\textbf{SRC artificial}} & \multicolumn{1}{c}{\textbf{TRG artificial}} & \multicolumn{1}{c}{\textbf{train}} & \multicolumn{1}{c}{\textbf{test}} & \multicolumn{1}{c}{\textbf{aug\_test}}  \\ 
        \cmidrule(lr){2-2}  \cmidrule(lr){3-3} \cmidrule(lr){4-4} \cmidrule(lr){5-5} \cmidrule(lr){6-6} \cmidrule(lr){7-7} \cmidrule(lr){8-8}
        \#1 & in [\smalldots] \texttt{NOUN} & in [\smalldots] \texttt{NOUN} & huheke [\smalldots] \texttt{NOUN} & \texttt{N}$+$jetah$+$\texttt{OUN} & 116 868 & 474 & 700\\
        \#2 & auf [\smalldots] \texttt{NOUN} & on [\smalldots] \texttt{NOUN} & siye [\smalldots] \texttt{NOUN} & \texttt{N}$+$dezaxe$+$\texttt{OUN} & 47 102 & 248 & 700\\
        \#3 & gegen [\smalldots] \texttt{NOUN} & against [\smalldots] \texttt{NOUN} & \texttt{N}$+$yusid$+$\texttt{OUN} & huxi  [\smalldots] \texttt{NOUN} & 15 540 & 137 & 600\\
        \#4 & bei [\smalldots] \texttt{NOUN} & at [\smalldots] \texttt{NOUN} & \texttt{N}$+$yadey$+$\texttt{OUN} & numime  [\smalldots] \texttt{NOUN} & 3796 & 92 & 600\\ \addlinespace \addlinespace
        & \multicolumn{7}{c}{\textbf{Vowel Harmony}} \\
         \cmidrule(lr){2-8} 
        & \multicolumn{1}{c}{\textbf{SRC pattern}} & \multicolumn{1}{c}{\textbf{TRG pattern}} & \multicolumn{1}{c}{\textbf{SRC artificial}} & \multicolumn{1}{c}{\textbf{TRG artificial}} & \multicolumn{1}{c}{\textbf{train}} & \multicolumn{1}{c}{\textbf{test}} & \multicolumn{1}{c}{\textbf{aug\_test}}  \\ 
        \cmidrule(lr){2-2}  \cmidrule(lr){3-3} \cmidrule(lr){4-4} \cmidrule(lr){5-5} \cmidrule(lr){6-6} \cmidrule(lr){7-7} \cmidrule(lr){8-8}
        \#1 & mit [\smalldots] \texttt{NOUN} & with [\smalldots] \texttt{NOUN} & duji  [\smalldots] \texttt{NOUN} & \texttt{NOUN} s-f-p & 133 082 & 679 & 700\\
        \#2 & zwei [\smalldots] \texttt{NOUN} & two [\smalldots] \texttt{NOUN} & zoged  [\smalldots] \texttt{NOUN} & \texttt{NOUN} b-p-r & 29 048 & 323 & 700\\
        \#3 & nach [\smalldots] \texttt{NOUN} & after [\smalldots] \texttt{NOUN} & \texttt{NOUN} n-l-j & dulana  [\smalldots] \texttt{NOUN} & 8636 & 174 & 700\\
        \#4 & vor [\smalldots] \texttt{NOUN} & before [\smalldots] \texttt{NOUN} & xefoqi  [\smalldots] \texttt{NOUN} & \texttt{NOUN} b-k-m & 7037 & 115 & 700\\ \addlinespace \addlinespace
        & \multicolumn{7}{c}{\textbf{Reduplication}} \\
         \cmidrule(lr){2-8} 
        & \multicolumn{1}{c}{\textbf{SRC pattern}} & \multicolumn{1}{c}{\textbf{TRG pattern}} & \multicolumn{1}{c}{\textbf{SRC artificial}} & \multicolumn{1}{c}{\textbf{TRG artificial}} & \multicolumn{1}{c}{\textbf{train}} & \multicolumn{1}{c}{\textbf{test}} & \multicolumn{1}{c}{\textbf{aug\_test}}  \\ 
        \cmidrule(lr){2-2}  \cmidrule(lr){3-3} \cmidrule(lr){4-4} \cmidrule(lr){5-5} \cmidrule(lr){6-6} \cmidrule(lr){7-7} \cmidrule(lr){8-8}
        Partial & sehr \texttt{ADJE} & very \texttt{ADJE} & \texttt{ADJ}$+$\texttt{ADJE} & popera & 34 783 & 121 & 700\\
        Triple & sehr, sehr \texttt{ADJE} & very, very \texttt{ADJE} & \texttt{ADJ}$+$\texttt{ADJ}$+$\texttt{ADJE} & metuza & 106 & 121 & 700\\
        Full & nicht \texttt{ADJE} & not \texttt{ADJE} & gija & \texttt{ADJE}$+$\texttt{ADJE} & 9664 & 50 & 563\\
    \end{tabular}
    \caption{Overview of pattern pairs for every morphological phenomenon, the artificial morphemes that we replace them with and how often they occur in the training data, the original test set (main results) and the augmented test set (results in Appendix \ref{sec:appendixc}).}
    \label{tab:data_overview}
\end{table*}

\subsection{Morphological Phenomena and Pattern Pairs}
\label{sec:appendixb}

Table \ref{tab:data_overview} shows the pattern pairs that we define for each morphological phenomenon. The first column shows what the pattern matches in the original source sentence, i.e.\ a random noun, a preposition, a cardinal number or a modifier of an adjective. The second column shows the corresponding pattern that we match in the target sentence. We always check that these patterns and the nouns or adjective following them are aligned and we make sure that the dependency relationships between them are correct, e.g.\ that the following noun is the head of the prepositional phrase.

Column three shows the artificial morphemes we use on the source side and column four the ones we use on the target side. For compounding, the morphological phenomenon always occurs on the source side. We simply concatenate the artificial morpheme with the matched noun. On the target side, we insert the artificial morpheme as a separate token before the noun. 

For the remaining morphological phenomena, we have patterns where the phenomenon occurs on the source side and others where they occur on the target side. Circumfixes are formed by deleting the matched preposition and adding an artificial morpheme before and after the noun. In the other sentence, we simply replace the preposition with an artificial morpheme. Infixes are formed in the same way but instead of adding the artificial morphemes before and after the noun, we insert one before the first vowel inside the noun.

We form the vowel harmony by deleting the preposition or cardinal number and inserting an artificial morpheme as a separate token after the noun. This morpheme is a placeholder consisting of three consonants. We then fill the positions between the consonants with the last two vowels occurring in the noun. If the noun only has one vowel, we insert this vowel twice. In the other sentence, we replace the preposition or cardinal number again with the artificial morpheme.

For partial reduplication, we extract a substring of the matched adjective until after the first vowel. We then repeat this substring to form a partially reduplicated adjective. If the adjective starts with a vowel, we extract the substring until after the second vowel. For triplication, we extract the same substring but repeat it twice and for full reduplication, we repeat the whole adjective. In the other sentence, we simply replace the modifier with an artificial morpheme.

For our abstract representation of morphological phenomena (not shown in the table), we generate a token of the form @TYPE\_\#@ that is inserted after the noun. For the other sentences, we have a distinct set of artificial morphemes that we use in the same way as for the surface form. Examples for the abstract representations can be seen in Table \ref{tab:synthetic}.

In column five and six, we present the frequencies with which these pattern pairs occur in the training data and the test set. We note that some pattern pairs are not as frequent as others and result in a relatively small test set, e.g.\ full reduplication. To check whether this affects our results, we present additional experiments on a synthetically balanced test set that we generate using data augmentation. We explain this further in Appendix \ref{sec:appendixc}. The pattern frequencies in this augmented test set are shown in the last column.

\subsection{Experiments on Synthetically Balanced Data}
\label{sec:appendixd}
We find that some pattern pairs do not occur very frequently in our training set (see Table \ref{tab:data_overview}) which raises the question of whether our results are meaningful enough. Furthermore, our test set may consist of more sentences for words with synthetic phenomena that we have seen a handful of times in the training data but fewer ``zero-shot'' cases. To check that our results are still valid, we create a synthetic, balanced test set using data augmentation and show that the results on this augmented test set are in line with the results presented in the main body of the paper.

To enrich our test set with synthetically generated sentence pairs for the evaluation, we use a very simple data augmentation technique. Specifically, we a) substitute prepositions with other prepositions, b) substitute numbers and other cardinals with the cardinal ``two'' and c) insert different modifiers before any adjectives. Below are some English example sentences after data augmentation:

\begin{center}
\begin{tabular}{c}
\vspace{0.2cm}
\textbf{Prepositions:} \\
\texttt{orig}: They came to Verona \textbf{from} Bologna.\\
\texttt{0.3440}: They came to Verona \textbf{with} Bologna.\\
\texttt{7.6578}: They came to Verona \textbf{against} Bologna.\\
\\
\textbf{Cardinals:} \\
\texttt{orig}: "I'm known to work \textbf{20} hours a day."\\
\texttt{-0.8844}: "I'm known to work \textbf{two} hours a day."\\
\\
\textbf{Intensity Markers:} \\
\texttt{orig}: The selection is broad.\\
\texttt{-0.6354}: The selection is \textbf{very} broad.\\
\texttt{1.8426}: The selection is \textbf{not} broad.\\
\vspace{0.2cm}
\end{tabular}
\end{center}

We score our synthetic test data with a language model and compute the difference in (pseudo)-perplexity to the original sentences to obtain a synthetic data score (see examples above - the lower the score the better). The scores for the German sentence and the English sentence are averaged to obtain a single score. The sentences are then ordered by this score such that we can pick the X most natural sentences for the evaluation. We use Masked-Language-Model-Scoring \citep{salazar-etal-2020-masked} and score both the German and the English sentences with the multilingual BERT model \citep{devlin-etal-2019-bert}.

We then define a set of seven frequency classes that capture how often a specific word has been seen with the morphological phenomenon in the training data: zero-shot, one to five times, six to 15 times, 16 to 50 times, 51 to 100 times, 101 to 500 times and 501 to 1000 times. For each of these buckets, we extract up to 100 sentences from the concatenated original and augmented test data. The original sentences are picked first and if necessary, we fill up each bucket with augmented sentences ordered by the language model score.

The results on the synthetically balanced test can be seen in Table \ref{tab:coarse_results_data_augmentation}. The results are very similar to the results on the original test sentences presented in Table \ref{tab:coarse_results}. Consequently, we conclude that the results presented in the main body of the paper are not affected by the imbalanced test data.

\begin{table*}[t]
    \centering
    \begin{tabular}{crcS[table-number-alignment = right]cccccccc}
    & & &  & \multicolumn{4}{c}{\textbf{Surface Representation}} & \textbf{Abstract}\\
    \cmidrule(lr){5-8} \cmidrule(lr){9-9} 
    &  & \textbf{Side} & \textbf{Train Freq.} &bpe32k & bpe-d32k & bpe-d500  & char & bpe32k\\
        \cmidrule(lr){3-3}   \cmidrule(lr){4-4}  \cmidrule(lr){5-5}  \cmidrule(lr){6-6}  \cmidrule(lr){7-7}  \cmidrule(lr){8-8} \cmidrule(lr){9-9} \addlinespace
    \multirow{5}{*}{\textbf{Compounding}} & \#9 & src & 27 & \phantom{0}0.0 & \phantom{0}0.0 & \phantom{0}0.0 & \phantom{0}0.0 & \phantom{0}0.0\\   
    & \#7 & src & 67 & \colorbox[HTML]{F17779}{44.8} & \phantom{0}0.0 & \colorbox[HTML]{F9CBD0}{\textbf{83.8}} & \phantom{0}0.0 & 95.5\\
    & \#5 & src & 238 & \colorbox[HTML]{5CC35C}{\textbf{98.3}} & \colorbox[HTML]{5CC35C}{97.8} & \colorbox[HTML]{5CC35C}{96.3} & \colorbox[HTML]{5CC35C}{97.0} & 98.3\\
    & \#3 & src & 522 & \colorbox[HTML]{5CC35C}{\textbf{99.0}} & \colorbox[HTML]{5CC35C}{98.3} & \colorbox[HTML]{5CC35C}{97.5} & \colorbox[HTML]{5CC35C}{96.8} & 98.3 \\
    & \#1 & src & 1095 & \colorbox[HTML]{5CC35C}{96.5} & \colorbox[HTML]{5CC35C}{\textbf{98.0}} & \colorbox[HTML]{5CC35C}{97.5} & \colorbox[HTML]{5CC35C}{97.3} & 96.8 \\ \addlinespace \addlinespace\multirow{4}{*}{\textbf{Circumfixation}}   & \#4 & src & 11718 & \colorbox[HTML]{5CC35C}{98.7} & \colorbox[HTML]{5CC35C}{\textbf{99.0}} & \colorbox[HTML]{5CC35C}{98.3} & \colorbox[HTML]{5CC35C}{95.7} & 98.8\\
    & \#2 & src & 26007 & \colorbox[HTML]{5CC35C}{\textbf{99.7}} & \colorbox[HTML]{5CC35C}{98.8} & \colorbox[HTML]{5CC35C}{99.0} & \colorbox[HTML]{5CC35C}{98.8} & 99.5 \\ \addlinespace
    & \#3 & trg & 21372 & \colorbox[HTML]{5CC35C}{97.7} & \colorbox[HTML]{5CC35C}{\textbf{99.1}} & \colorbox[HTML]{5CC35C}{98.6} & \colorbox[HTML]{5CC35C}{97.7} & 99.7\\ 
   & \#1 & trg & 122017 & \colorbox[HTML]{5CC35C}{96.4} & \colorbox[HTML]{5CC35C}{\textbf{99.3}} & \colorbox[HTML]{5CC35C}{99.1} & \colorbox[HTML]{5CC35C}{97.7} & 99.6 \\ 
    \addlinespace \addlinespace \multirow{4}{*}{\textbf{Infixation}} & \#4 & src & 3796 & \colorbox[HTML]{5CC35C}{\textbf{99.2}} & \colorbox[HTML]{5CC35C}{98.8} & \colorbox[HTML]{5CC35C}{97.3} & \colorbox[HTML]{5CC35C}{98.0} & 99.3 \\ 
    & \#3 & src & 15540 & \colorbox[HTML]{5CC35C}{98.7} & \colorbox[HTML]{5CC35C}{96.8} & \colorbox[HTML]{5CC35C}{\textbf{99.3}} & \colorbox[HTML]{5CC35C}{98.7} & 98.5 \\ \addlinespace
    & \#2 & trg & 47102 & \colorbox[HTML]{5CC35C}{97.1} & \colorbox[HTML]{5CC35C}{\textbf{98.4}} & \colorbox[HTML]{5CC35C}{98.3} & \colorbox[HTML]{5CC35C}{97.3} & 99.7\\ 
    & \#1 & trg & 116868 & \colorbox[HTML]{5CC35C}{97.4} & \colorbox[HTML]{5CC35C}{\textbf{98.3}} & \colorbox[HTML]{5CC35C}{97.9} & \colorbox[HTML]{5CC35C}{98.1} & 99.7\\ 
    \addlinespace \addlinespace
    \multirow{4}{*}{\textbf{Vowel Harmony}} & \#3 & src & 8636 & \colorbox[HTML]{5CC35C}{\textbf{99.0}} & \colorbox[HTML]{5CC35C}{98.0} & \colorbox[HTML]{5CC35C}{96.7} & \colorbox[HTML]{5CC35C}{97.1} & 98.9\\ \addlinespace
    & \#4 & trg & 7037 & \colorbox[HTML]{F17779}{57.3} & \colorbox[HTML]{F17779}{73.0} & \colorbox[HTML]{B2EAB1}{92.0} & \colorbox[HTML]{B2EAB1}{\textbf{92.6}} & 99.9 \\
    & \#2 & trg & 29048 & \colorbox[HTML]{F17779}{76.1} & \colorbox[HTML]{F9CBD0}{81.4} & \colorbox[HTML]{B2EAB1}{\textbf{93.6}} & \colorbox[HTML]{B2EAB1}{92.3} & 99.7\\ 
    & \#1 & trg & 133082 & \colorbox[HTML]{F9CBD0}{82.6} & \colorbox[HTML]{F9CBD0}{88.3} & \colorbox[HTML]{B2EAB1}{94.0} & \colorbox[HTML]{B2EAB1}{\textbf{94.6}} &  99.4\\  \addlinespace \addlinespace
    \multirow{3}{*}{\textbf{Reduplication}}& Triple & src & 106 & \phantom{0}0.0	& \phantom{0}0.0 & \phantom{0}0.0 & \phantom{0}0.0 & 98.6 \\ 
     & Partial & src & 34783 &  \colorbox[HTML]{F9CBD0}{84.1} & \colorbox[HTML]{B2EAB1}{\textbf{90.7}} & \colorbox[HTML]{F9CBD0}{89.1} & \colorbox[HTML]{F9CBD0}{84.4} & 98.6 \\ \addlinespace
    & Full & trg & 9664 & \colorbox[HTML]{F17779}{74.2} & \colorbox[HTML]{F9CBD0}{88.6} & \colorbox[HTML]{B2EAB1}{\textbf{93.4}}  & \colorbox[HTML]{B2EAB1}{91.8} & 99.1
    \end{tabular}
    \caption{Accuracy (in \%) of the four models for each of the morphological pattern pairs \textbf{on the augmented test set}. Best results for surface representation  are marked in bold. $\geq$95\% dark green, $\geq$90\% light green,  $\geq$80\% light red,  $<$80\% dark red (best viewed in colour). Patterns ordered by src / trg side, then by frequency.} 
    \label{tab:coarse_results_data_augmentation}
\end{table*}

\subsection{Error Analysis}
\label{sec:appendixc}
 We manually check up to 50 incorrect translations per pattern pair in the original test set. For classification of the errors, we define the following error types:

\begin{itemize}
    \item[\texttt{M1}] no artificial morpheme in output
    \item[\texttt{S1}] source artificial morpheme, base untranslated
    \item[\texttt{S2}] only source artificial morpheme untranslated
    \item[\texttt{S3}] source artificial morpheme translated to orthographically similar word
    \item[\texttt{T1}] target artificial morpheme not entirely correct (e.g.\ wrong vowels in vowel harmony)
    \item[\texttt{T2}] target artificial morpheme translated as orthographically similar word
    \item[\texttt{T3}] target artificial morpheme occurs multiple times
    \item[\texttt{T4}] word break between artificial morpheme $+$ base
    \item[\texttt{T5}] concatenation with semantically similar word instead of reduplication
    \item[\texttt{O1}] other, unrelated artificial morpheme generated
    \item[\texttt{A1}] abstract instead of surface form generated
\end{itemize}

We summarise the most interesting findings in Section \ref{sec:error_types} and list the full distribution of error types for the pattern pairs here:\\\\\\

\textbf{Higher-Resource Compounds (\#1, \#3, \#5):} \\
\begin{tabular}{lrrrr} \addlinespace
     & bpe32k & bpe-d32k & bpe-d500 & char\\ \addlinespace
    \texttt{M1} & 60\% & 87\% & 91\% & 97\%\\\addlinespace
    \texttt{S1} & 24\% &  -   &  -   &  -\\\addlinespace
    \texttt{S2} &  4\% &  -   &  -   &  3\%\\\addlinespace
    \texttt{S3} &  4\% &  -   &  -   &  -\\ \addlinespace
    \texttt{T3} &  8\% & 13\% &  9\% &  -\\\addlinespace
\end{tabular}\\\\
\newpage
\textbf{Lower-Resource Compounds (\#7, \#9):} \\
\begin{tabular}{lrrrr} \addlinespace
     & bpe32k & bpe-d32k & bpe-d500 & char\\ \addlinespace
    \texttt{M1} & 18\% &  6\% & 13\% & 12\%\\\addlinespace
    \texttt{S1} & 16\% &  7\% & 14\% & 11\%\\\addlinespace
    \texttt{S2} & 47\% & 87\% & 44\% & 55\%\\\addlinespace
    \texttt{S3} & 10\% &  -   &  4\% &  8\%\\ \addlinespace
    \texttt{T2} &  -   &  -   & 25\% & 14\%\\ \addlinespace
    \texttt{O1} &  9\% &  -   &  -   &  -\\ 
\end{tabular}\\\\

\textbf{Circumfixation on Source Side (\#2, \#4):} \\
\begin{tabular}{lrrrr} \addlinespace
     & bpe32k & bpe-d32k & bpe-d500 & char\\ \addlinespace
    \texttt{M1} & 100\% &  86\% & 71\% & 100\%\\\addlinespace
    \texttt{T3} &   -   &  14\% & 29\% &  -\\ 
\end{tabular}\\\\\\

\textbf{Circumfixation on Target Side (\#1, \#3):} \\
\begin{tabular}{lrrrr} \addlinespace
     & bpe32k & bpe-d32k & bpe-d500 & char\\ \addlinespace
    \texttt{M1} & 19\% &  6\% & 13\% & 12\%\\\addlinespace
    \texttt{T1} & 38\% &  7\% & 14\% & 11\%\\\addlinespace
    \texttt{T4} & 43\% & 87\% & 44\% & 55\%\\
\end{tabular}\\\\\\

\textbf{Infixation on Source Side (\#3, \#4):} \\
\begin{tabular}{lrrrr} \addlinespace
     & bpe32k & bpe-d32k & bpe-d500 & char\\ \addlinespace
    \texttt{M1} & 100\% & 83\% & 75\% & 100\%\\\addlinespace
    \texttt{T3} & -    &  17\% & 25\% & -\\
\end{tabular}\\\\\\

\textbf{Infixation on Target Side (\#1, \#2):} \\
\begin{tabular}{lrrrr} \addlinespace
     & bpe32k & bpe-d32k & bpe-d500 & char\\ \addlinespace
    \texttt{M1} & 94\% &  89\% & 100\% & 100\%\\\addlinespace
    \texttt{T3} & -    &  11\% & -     & -\\\addlinespace
    \texttt{A1} &  6\% &   -   & -     & -\\
\end{tabular}\\\\\\

\textbf{Vowel Harmony on Source Side (\#3):} \\
\begin{tabular}{lrrrr} \addlinespace
     & bpe32k & bpe-d32k & bpe-d500 & char\\ \addlinespace
    \texttt{M1} & 100\% &  100\% & 100\% & 100\%\\
\end{tabular}\\\\\\

\textbf{Vowel Harmony on Target Side (\#1, \#2, \#4):} \\
\begin{tabular}{lrrrr} \addlinespace
     & bpe32k & bpe-d32k & bpe-d500 & char\\ \addlinespace
    \texttt{T1} & 100\% &  100\% & 100\% & 100\%\\
\end{tabular}\\\\\\

\textbf{Partial Reduplication on Source Side:} \\
\begin{tabular}{lrrrr} \addlinespace
     & bpe32k & bpe-d32k & bpe-d500 & char\\ \addlinespace
    \texttt{M1} & 100\% &  100\% & 100\% & 100\%\\
\end{tabular}\\\\\\

\textbf{Triplication on Source Side:} \\
\begin{tabular}{lrrrr} \addlinespace
     & bpe32k & bpe-d32k & bpe-d500 & char\\ \addlinespace
    \texttt{M1} & 6\% &  6\% & 4\% & 2\%\\ \addlinespace
    \texttt{O1} & 94\% &  94\% & 96\% & 98\%\\ 
\end{tabular}\\\\\\

\textbf{Full Reduplication on Target Side:} \\
\begin{tabular}{lrrrr} \addlinespace
     & bpe32k & bpe-d32k & bpe-d500 & char\\ \addlinespace
    \texttt{M1} & 7\% &  25\% & 100\% & 40\%\\ \addlinespace
    \texttt{T1} & - &  - & - & 60\%\\  \addlinespace
    \texttt{T4} & 14\% &  12\% & - & -\\  \addlinespace
    \texttt{T5} & 79\% &  63\% & - & -\\  
\end{tabular}\\\\\\

\end{document}